\documentclass{Interspeech2024}

\usepackage{multirow}
\usepackage{xspace}
\usepackage{siunitx}
\usepackage{algorithm}
\usepackage{algpseudocode}
\usepackage{hyperref}
\usepackage{svg}

\def\F0{$F_0$\xspace}

\interspeechcameraready 

\title{A Human-in-the-Loop Approach to Improving Cross-Text Prosody Transfer}

\name[affiliation={1*}]{Himanshu}{Maurya}
\name[affiliation={1*}]{Atli}{Sigurgeirsson}

\address{
  $^1$The Centre for Speech Technology Research, University of Edinburgh, United Kingdom}
  
\email{lordzuko.research@gmail.com, s2063518@ed.ac.uk}


\begin{document}

\maketitle

\begin{abstract}
    \noindent Text-To-Speech (TTS) prosody transfer models can generate varied prosodic renditions, for the same text, by conditioning on a reference utterance. These models are trained with a reference that is identical to the target utterance. But when the reference utterance differs from the target text, as in cross-text prosody transfer, these models struggle to separate prosody from text, resulting in reduced perceived naturalness.

    To address this, we propose a Human-in-the-Loop (HitL) approach. HitL users adjust salient correlates of prosody to make the prosody more appropriate for the target text, while maintaining the overall reference prosodic effect. Human adjusted renditions maintain the reference prosody while being rated as more appropriate for the target text $57.8\%$ of the time. Our analysis suggests that limited user effort suffices for these improvements, and that closeness in the latent reference space is not a reliable prosodic similarity metric for the cross-text condition.
    
\end{abstract}

\def\thefootnote{*}\footnotetext{These authors contributed equally to this work}\def\thefootnote{\arabic{footnote}}

\section{Introduction} \label{sec:intro}
Prosody is used to encode information that is not fully conveyed by linguistic information alone. The same lexical information can be encoded through different prosodic renditions to deliver different meanings and influence particular interpretations \cite{wilson2006relevance}. Knowing which rendition is most appropriate at any given time is a non-trivial problem. Typical end-to-end TTS models generate a single \textit{mean} prosodic rendition, reflecting the overall prosodic distribution of the training corpus. Treating prosody in this way yields speech that can be perceived as dull, inexpressive, or inappropriate for the given target text, which negatively affects general comprehension \cite{govender2018using}.

\begin{figure}[!h]
    \centering
    \includegraphics[width=0.7\linewidth]{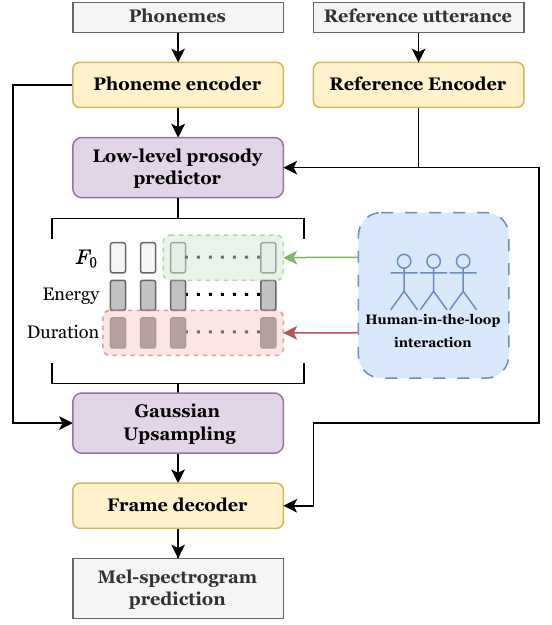}
    \caption{Overview of the proposed method. Human-in-the-loop participants adjust word- and utterance-level \F0, energy and duration features to improve prosody in cross-text prosody transfer.}
    \label{fig:model_overview}
\end{figure}

\section{Background} \label{sec:background}
One way of controlling prosody is to learn the prosodic distribution of the training corpus using a prosody-labelled TTS corpus. Prosody labels can be used to jointly train prosody predictors, which can be controlled to generate different prosodic renditions for the same input text \cite{Cai2020EmotionCS, Henter2018DeepEM, Wan2019CHiVEVP}. However, few prosody-labelled corpora are generally available. Therefore, much of the recent work on prosody control has shifted to methods that aim to discover the prosodic distribution without supervision. This includes Prosody-Transfer (PT) models \cite[e.g.]{pmlr-v80-skerry-ryan18a, Zadi2021DaftExprtRP, karlapati2020copycat}.

Typically, PT models jointly train a \textit{reference encoder} with an underlying acoustic model. This encoder models a highly constrained representation of the target acoustics, which is taken to represent prosody. The prosody representation is then used to condition speech generation in the underlying acoustic model. Under the right conditions, such models are claimed to be able to transfer prosody from any reference to any text \cite{Zadi2021DaftExprtRP}. 

A common PT use case is \textit{cross-text} PT; a reference with the target prosody exists, but the target text does not match the reference text. Under cross-text conditions, however, some PT models have demonstrated a trade-off between perceived naturalness and the overall quality of prosody transfer \cite{Sigurgeirsson2023DoPT}. This effect is commonly attributed to \textit{acoustic feature entanglement} \cite{pmlr-v80-skerry-ryan18a} and \textit{source-speaker leakage} \cite{karlapati2020copycat}. The fact that PT models are typically not trained on cross-text or cross-speaker samples is believed to give rise to these issues \cite{Sigurgeirsson2023DoPT}. As a result of the PT training regime, the modelled prosody representations are highly dependent on the reference text, which means that they are not strictly \textit{transferable} to any text \cite{Sigurgeirsson2023DoPT}.

These issues with PT models do give rise to questions about whether cross-text prosody transfer can even be achieved \textit{faithfully}. That is, can you transfer prosody to any text such that the resulting rendition is true to the reference prosody \textit{and} is perceived natural for the target text? If the latent reference space models transferable representations of prosody, then a prosody representation that fits that goal should exist. But if it is not retrievable by conditioning on the reference prosody, it becomes a question about how else that representation can be found in the reference space.

One possible way to do this would be to directly incorporate human perception in prosody prediction, using feedback from a \textit{Human-in-the-loop} (HitL) \cite{hil-survey}. Human feedback has been used for a wide spectrum of tasks in Natural Language Processing (NLP); such as entity extraction and linking \cite{gentile2019explore, zhang2019invest, klie2020zero} and more subjective tasks such as reading comprehension \cite{bartolo2020beat} and question-answering \cite{wallace2019trick}. Fine-tuning models on relatively small sets of human-annotated corrections in these tasks has shown improved model performance and robustness \cite{hil-survey}. HitL has also been employed in several TTS-specific tasks, such as modelling emotion \cite{gibbs_sampling}, speaker identity  \cite{Udagawa2022HumanintheloopSA}, and speaking style \cite{Cornille2022InteractiveMP}. These attempts demonstrate that HitL can, to some extent, be applied to TTS to improve particular perceptual objectives. 

An important consideration for any HitL based method is how participants interact with the underlying model. Previous work such as \cite{gibbs_sampling, Rijn2021ExploringEP} use HitL to explore emotional prosodic variation of neutral utterances. In their work, HitL participants make iterative adjustments in isolated dimensions of a learnt latent prosody space to explore different variations of emotional prosody. HitL interaction with the model is shown to make improvements compared to the baseline model. However, it is a very time-consuming approach, requiring 48 hours to generate 40 samples in \cite{Rijn2021ExploringEP}. Furthermore, applying HitL directly in a modelled latent space could be highly uninterpretable, as latent dimensions are not guaranteed to encode any perceptually distinct information type or be independent of other latent dimensions \cite{Mohan2021CtrlPTC}. Some methods \cite{Mohan2021CtrlPTC, Raitio2020ControllableNT} instead, control prosody in terms of known acoustic correlates of prosody, such as \F0, duration, spectral tilt and energy. There is no established \textit{control-resolution} for HitL based prosody control; \cite{Raitio2020ControllableNT} suggests using utterance-level control, while \cite{Mohan2021CtrlPTC} controls prosody at the phone-level. Although \cite{Mohan2021CtrlPTC} suggests that phone-level control allows HitL participants to make detailed decisions to improve the output prosody, they also suggest that a higher-level abstraction is perhaps more conducive for a HitL approach.

We investigate whether the prosody representations modelled by a particular PT-model \cite{Zadi2021DaftExprtRP} can be improved specifically for the cross-text PT task. The PT model separately models phone-level \F0, energy and duration. These are salient acoustic correlates of prosody, and together they cover the \textit{control space} that our HitL participants interact with. Participants control utterance-level and word-level features through a web-based user interface designed to carry out the experiments. The proposed design enables fewer, interpretable edit iterations when compared to prior HitL work in TTS. We evaluate the human effort involved in achieving the desired prosodic effect for the cross-text PT task. Our results indicate that HitL participants can discover prosodic renditions that are more appropriate for the given target text while maintaining most of the reference prosody effect.

\section{Method} \label{sec:method}
\subsection{Model Architecture} 
The proposed HitL method does not require any online supervision and can therefore be implemented using any pre-trained controllable PT-model. Our experiments aim to provide insights into the wider PT-task, instead of any specific model. We therefore choose \textit{Daft-Exprt} \cite{Zadi2021DaftExprtRP} as our baseline, since they demonstrate good performance in the cross-text setting. Daft-Exprt is a fully parallel TTS model that builds on the FastSpeech-2 architecture. Daft-Exprt uses a reference-encoder which generates a fixed-size reference embedding. This embedding is used to condition speech generation in the underlying acoustic model to enable prosody transfer. Conditioning is performed by predicting FiLM-parameters \cite{Perez2017FiLMVR} for the \textit{low-level prosody predictor} and the frame decoder. The low-level prosody predictor predicts phone-level log- \F0, energy and duration. These predicted values, with encoded phonemes, are used to decode the target Mel-spectrogram. We use HiFi-GAN \cite{Kong2020HiFiGANGA} to convert the predicted Mel-spectrograms to waveform.

\subsection{Human-in-the-loop approach}
HitL participants interact with Daft-Exprt through the low-level prosody predictor. Given a reference and a target text, an initial cross-text PT sample is generated. Participants are asked to adjust the predicted \F0, energy and duration values to make the prosody more appropriate for the target text, while preserving the prosody of the reference. Manipulating individual phone-level predictions is believed to be too complex for the proposed HitL task \cite{Mohan2021CtrlPTC}. So instead, we devise a control interface for making utterance- and word-level adjustments to prosody. Taking inspiration from existing interfaces for interaction with TTS systems \cite{9023345, TITS2021100055}, we develop a web-based User Interface (UI), using Streamlit\footnote{https://streamlit.io/}, to facilitate the proposed HitL method. Control inputs are received through the UI and corresponding \F0, energy and duration are computed, as explained in Sections \ref{subsec:local_prosody_control}-\ref{subsec:global_prosody_control}. These values are then sent to the acoustic model to complete synthesis.

\subsubsection{Word-level Control}
\label{subsec:local_prosody_control}
Daft-Exprt predicts phone-level \F0, energy and duration. Therefore, word-level control inputs have to be mapped to corresponding phone-level adjustments. We treat \F0 and energy adjustments in the same way, while treating duration adjustments slightly differently. First, the acoustic model predicts all phone-level \F0 and energy values $v_1, v_2, \dots, v_n$ (representing either phone-level \F0 or energy predictions henceforth). We then compute a word-level feature mean, $K_w$, and a per-phone scaling factor, $s_i$ for each input word $w=p_1,p_2,...,p_n$:
\begin{align*}
    K_w=\frac{1}{|w|}\sum v_i, \quad s_i = \frac{K_w}{v_i}, \quad i\in{1,\dots,n}
\end{align*}
The user can then suggest changing the initial mean feature value $K_w$ to a new one, $K^{\prime}_w$. This results in per-phone changes of that word in such a way that:
\begin{align*}
    v^{\prime}_i = \frac{K^{\prime}_w}{s_i}, \quad i\in{1,\dots,n}
\end{align*}
This form of control therefore results in equal adjustments to all phones $p_i$ in the word, proportional to their originally predicted value $v_i$. Note that in case of \F0, voiceless phones are not modified and do not contribute to the computation of the value of $K_w$. This control mechanism is presented to the user in the form of a \textit{slider}. The control slider is initially set to $K_w$, allowing for separately increasing or decreasing \F0 and energy values for each word in the input text. \F0 and energy values far outside the known training distribution can result in unintelligible output. We therefore determine a suitable range for both \F0 and energy modifications. From our initial experimentation, we found that limiting the \F0 range to $\pm3\sigma$ and energy to $\pm1.5\sigma$ resulted in a good control range for this task. We therefore limit \F0 and energy control such that no edited phone-level \F0 or energy value falls outside this range.

This control mechanism is counterintuitive for duration, since, by its design, it would set the initial duration of every word to its phone-duration mean. Therefore, we resort to a simpler approach for duration control. Participants can equally scale the duration of each phone in the word by a constant within the range $[0, 2]$. That is, each word can be made up to twice as long in duration. This design technically allows participants to set the duration of any word to $0$ seconds, but no HitL participant did. Like with \F0 and energy, word-level duration is controlled using a slider UI element.

\subsubsection{Utterance-level control} \label{subsec:global_prosody_control}
We also provide utterance-level controls for \F0, energy and duration. Users also interact with these controls through slider UI elements. We believe that applying global changes to prosody may help participants better control emotion, expression, and overall tone of the output. This is similar to the control scheme used in \cite{sigurgeirsson2023using}. Again, we treat \F0 and energy control in the same way. When a HitL participant submits an utterance-level control input, we compute the appropriate word-level control input for each word and apply it. We determine the utterance-level control-range such that any resulting word-level change remains within the statistical per-phone ranges described in Section \ref{subsec:local_prosody_control}. The utterance-level duration control simply allows participants to scale all phone durations equally within the range $[0,2]$.

\section{Experiments} \label{sec:experiments}
\subsection{Baseline Model}
The baseline Daft-Exprt model is trained on the 2013 Blizzard Challenge corpus \cite{King2014TheBC}, which is an expressive single-speaker book narration corpus. We use only the segmented split of the corpus, which comprises 52 hours from 55 different books. Utterances shorter than $0.3$ seconds and longer than $15$ seconds are excluded. The remaining, approximately 40,000 utterances ($\approx40$ hours), are used to train the baseline PT-model. We follow the original Daft-Exprt preprocessing procedure \cite{Zadi2021DaftExprtRP} and extract alignments, using the Montreal forced aligner \cite{mcauliffe17_interspeech}, and phone-level \F0 and energy estimations.

The model is trained for 24 hours, distributed over 4 NVIDIA Tesla V100-SXM2-16GB GPUs, using a batch size of 96 samples. We train a HiFi-GAN \cite{Kong2020HiFiGANGA} vocoder using the same corpus as is used to train the acoustic model. The vocoder is trained for 12 hours on a single NVIDIA Tesla V100-SXM2-16GB GPU using a batch size of 8 samples.

\subsection{Edited Speech Corpus} \label{sec:edited_corpus}
In our proposed method, participants first listen to the reference utterance used for the cross-text PT. They are then asked to make adjustments to \F0, energy, and duration to make the resulting prosody more appropriate for the target text. Participants are asked to make these adjustments while maintaining the perceived naturalness of the output and the overall prosodic effect of the reference utterance. After making the adjustments, the \textit{edited} sample and the initially synthesised cross-text PT sample are saved. We also ask participants to indicate how confident they are (\textit{"low"}/\textit{"high"}) in their suggestions.

To account for any possible individual bias, we ask all of our participants to make adjustments to the same list of cross-text PT samples. We choose 5 reference utterances on the basis that each one would likely yield a perceptually distinct effect in the PT output. We create four target texts for the task that would typically elicit a particular prosodic effect. The target texts are all short (5 words or fewer) to keep the HitL process as simple as possible. As a result, the target texts are notably shorter than the text read for the reference utterances we use. The 5 references and 4 target texts create $5\times4=20$ reference-target pairs which are used to generate the 20 cross-text PT samples that each HitL participant modifies. A full list of references and target texts is available on our demo page\footnote{\hyperlink{https://linktr.ee/anonymousconferenceuser}{https://linktr.ee/anonymousconferenceuser}}.

\begin{figure}[!h]
    \centering
\includegraphics[width=\linewidth]{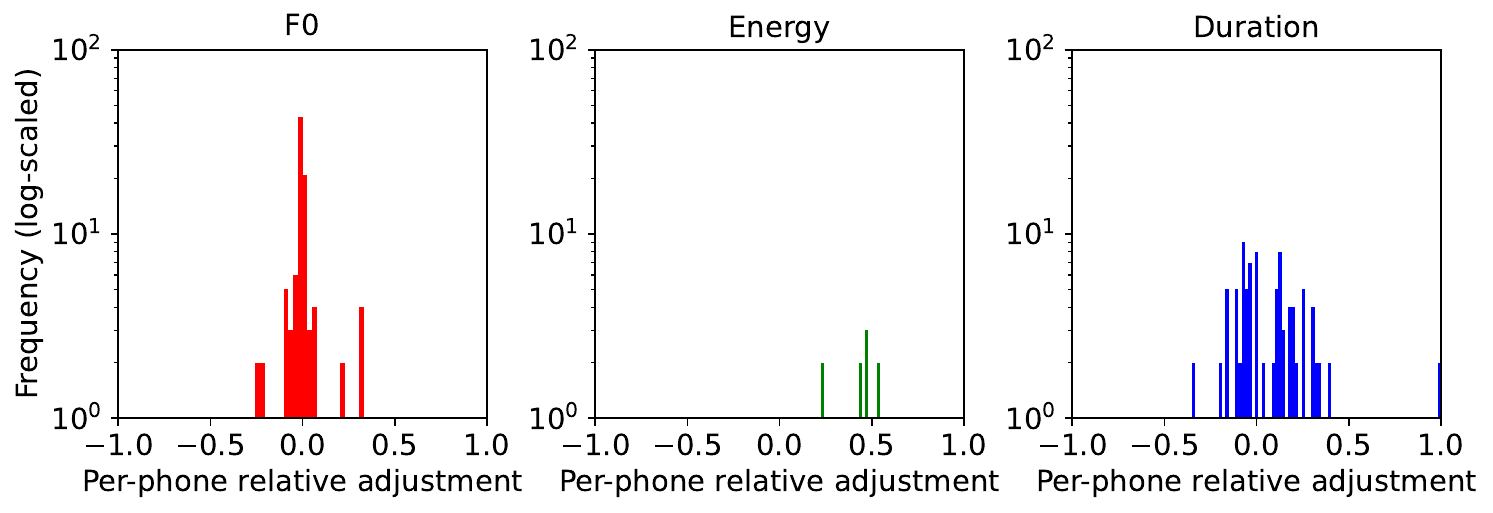}
    \caption{The distributions of phone-level control-inputs for \F0, energy and duration (unchanged values omitted). These results suggest that energy control was substantially less important to our HitL participants than \F0 and duration control.}
    \label{fig:relative_edits}
\end{figure}

We recruit 33 HitL participants with a diverse language background to take part in our study. Their participation resulted in $5\times4\times33=660$ \textit{original} / \textit{edited} sample pairs. Participants spent on average $123.5\pm138.0$ seconds modifying each sample, resulting in on average $7.0\pm7.1$ individual operations per utterance. Of the 660 pairs, 193 sample pairs were unmodified. In these cases, the participants deemed the \textit{original} prosodic rendition appropriate enough for the target text. These identical pairs and 47 additional unusable pairs were removed from the final set. The resulting 420 pairs are used in our subjective and objective experiments. Of the 420 pairs, the participants indicated a high confidence in their performance for $82.6\%$ and a low confidence for $17.4\%$. Interestingly, as shown in Figure \ref{fig:relative_edits}, participants made around $3\times$ fewer edits to predicted energy values when compared to \F0 and duration. This could mean that HitL participants did not find energy-control to be as useful or important to complete the task.

\subsection{Evaluation} \label{sec:evaluation}
We evaluate perceived naturalness, quality of prosody transfer, and prosodic appropriateness. We use a 5-point Likert scale to evaluate perceived naturalness in a standard Mean Opinion Score (\textbf{MOS}) survey. We follow the \textbf{MUSHRA-like} design proposed in \cite{pmlr-v80-skerry-ryan18a} to evaluate prosody transfer. We also include a sample that is generated using a random reference utterance. A random reference is unlikely to be informative about the target prosody, so we view this randomly generated sample as an anchor in our design. Participants first listen to the reference utterance used to generate the \textit{original} cross-text PT sample. Then they listen to the \textit{original} sample, the \textit{edited} sample, and the random anchor. They then indicate, on a scale from 0 to 100, how prosodically similar these samples are to the reference. We use a simple \textbf{A/B} preference design to evaluate the prosodic preference between \textit{original} and \textit{edited} samples. Participants are asked to indicate this preference in terms of how appropriate the prosodic rendition is for the synthesised text.

We evaluate all the 420 pairs described in Section \ref{sec:edited_corpus}. This results in 840 MOS questions, 420 MUSHRA-like screens, and 420 A/B preference questions. We recruited 68 native UK/US participants through Prolific\footnote{https://www.prolific.com/}. Each sample is evaluated by at least 3 different raters, and at most 5. Table \ref{tab:results} summarises our subjective evaluation results.

\begin{table}[!h]
\centering
\caption{Main subjective results broken down by HITL participant confidence. The highlighted row includes high-confidence results of the proposed method.} \label{tab:results}
\begin{tabular}{ccccc}
\multicolumn{2}{c}{\textbf{\begin{tabular}[c]{@{}c@{}}Participant\\ confidence\end{tabular}}} &
  \multicolumn{1}{c}{\textbf{MOS}} &
  \multicolumn{1}{c}{\textbf{\begin{tabular}[c]{@{}c@{}}MUSHRA\\ like\end{tabular}}} &
  \multicolumn{1}{c}{\textbf{A/B}} \\ \hline
\toprule
\multirow{2}{*}{Low}   & \textit{original} & $3.1\pm0.7$                              & $61.2\pm13.4$                              & $50.7\%$                              \\
                       & \textit{edited}   & $2.7\pm0.8$                              & $53.0\pm18.1$                              & $49.3\%$                              
\\ \midrule
\multirow{2}{*}{High}  & \textit{original} & \multicolumn{1}{l}{$3.2\pm0.7$}          & \multicolumn{1}{l}{$58.8\pm13.1$}          & \multicolumn{1}{l}{$40.4\%$}          \\
                       & \textbf{\textit{edited}}   & \multicolumn{1}{l}{$\mathbf{3.0\pm0.8}$} & \multicolumn{1}{l}{$\mathbf{55.0\pm14.3}$} & \multicolumn{1}{l}{$\mathbf{59.6\%}$} \\ \midrule
\multirow{2}{*}{Total} & \textit{original} & $3.2\pm0.7$                              & $59.2\pm13.2$                              & $42.1\%$                              \\
                       & \textit{edited}   & \textbf{$3.0\pm0.7$}                     & $54.7\pm13.2$                              & $57.8\%$                \\
                       \bottomrule
\end{tabular}
\end{table}

\noindent The \textit{edited} samples are perceived to be slight, although not significantly, less prosodically similar to the reference. The \textit{edited} samples are also perceived as slightly less natural than the \textit{original} samples. However, when HitL participants indicate high confidence in their efforts, this difference decreases. Furthermore, in these cases, the evaluators indicate a high preference for the \textit{edited} prosodic renditions at a rate of $59.6\%$. We hypothesised that an improvement in the prosodic renditions of cross-text PT samples would consequently translate into an improvement in perceived naturalness. Our results do not support that claim. However, we hypothesise that the drop in perceived naturalness can largely be explained by artefacts, audible discontinuities and interruptions introduced by our HitL control procedure.

\vspace{-2mm}
\begin{figure}[!h]
    \centering
\includegraphics[width=0.6\linewidth]{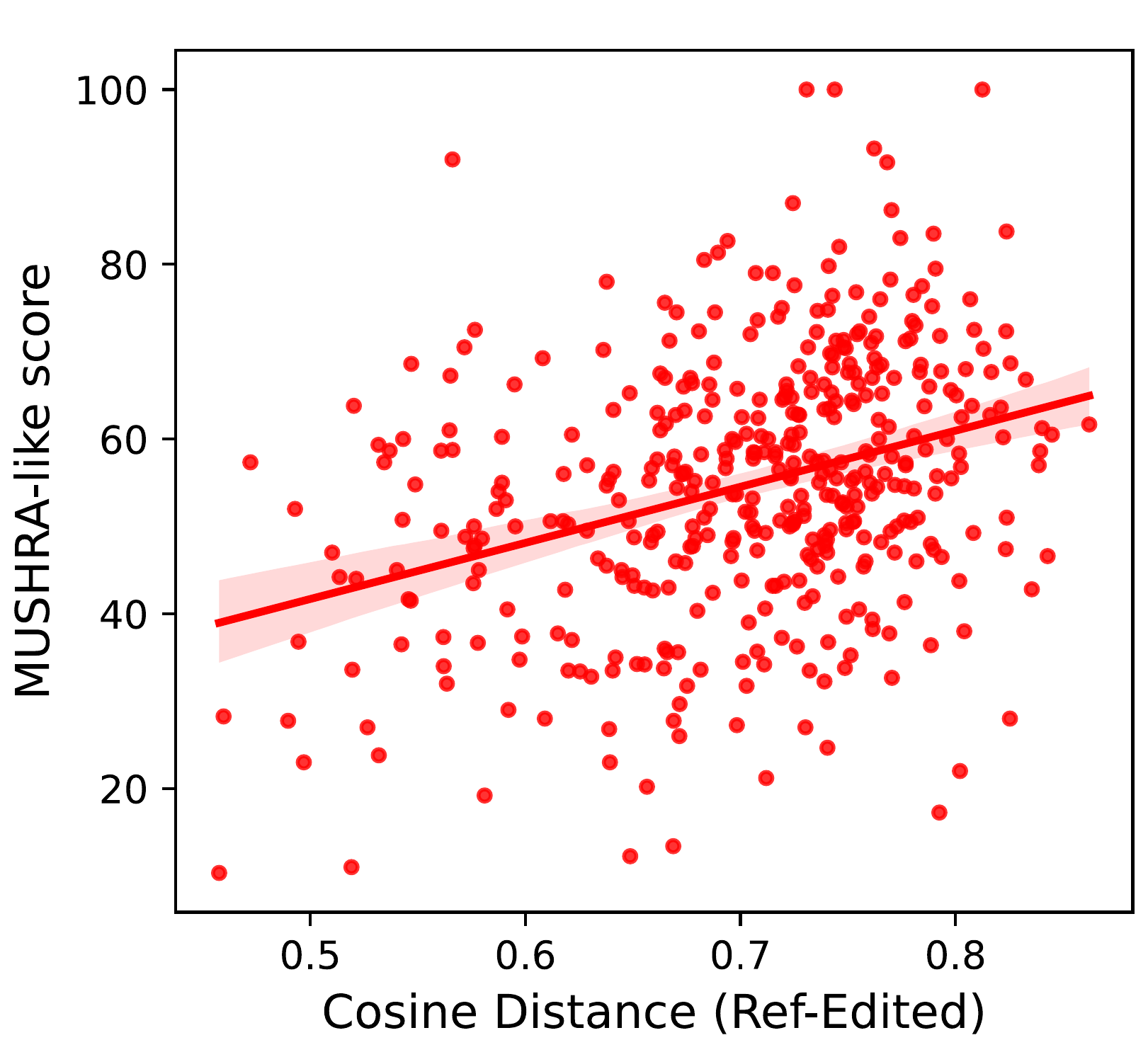}
    \caption{MUSHRA-like scores of edited cross-text PT samples plotted against the cosine distance between the original reference embedding and the embedded edited cross-text PT sample. A linear regression model is fitted to this data, the shaded red area indicates the $95\%$ confidence interval.}
    \label{fig:distance}
\end{figure}

\noindent We investigate the relationship between the reference embeddings and embedded synthesised samples. We speculated that synthesised PT samples that are rated as prosodically similar to the reference would be, once embedded, close to the reference embedding in the latent space. However, we observe the opposite as shown in Figure \ref{fig:distance}. Edited cross-text PT samples that are found to be prosodically similar to the reference tend to be further away from the reference embedding in the latent space. A similar trend was not found for unedited cross-text PT samples. We believe these results demonstrate two things: 1) participants can identify a\textit{``prosodic intent"} from the reference and \textit{faithfully} modify the cross-text PT sample with regard to the identified intent and the target text, and 2) Closeness to a reference embedding is not a reliable metric for prosodic similarity for cross-text PT.

\vspace{-2mm}
\begin{figure}[!h]
    \centering
\includegraphics[width=0.8\linewidth]{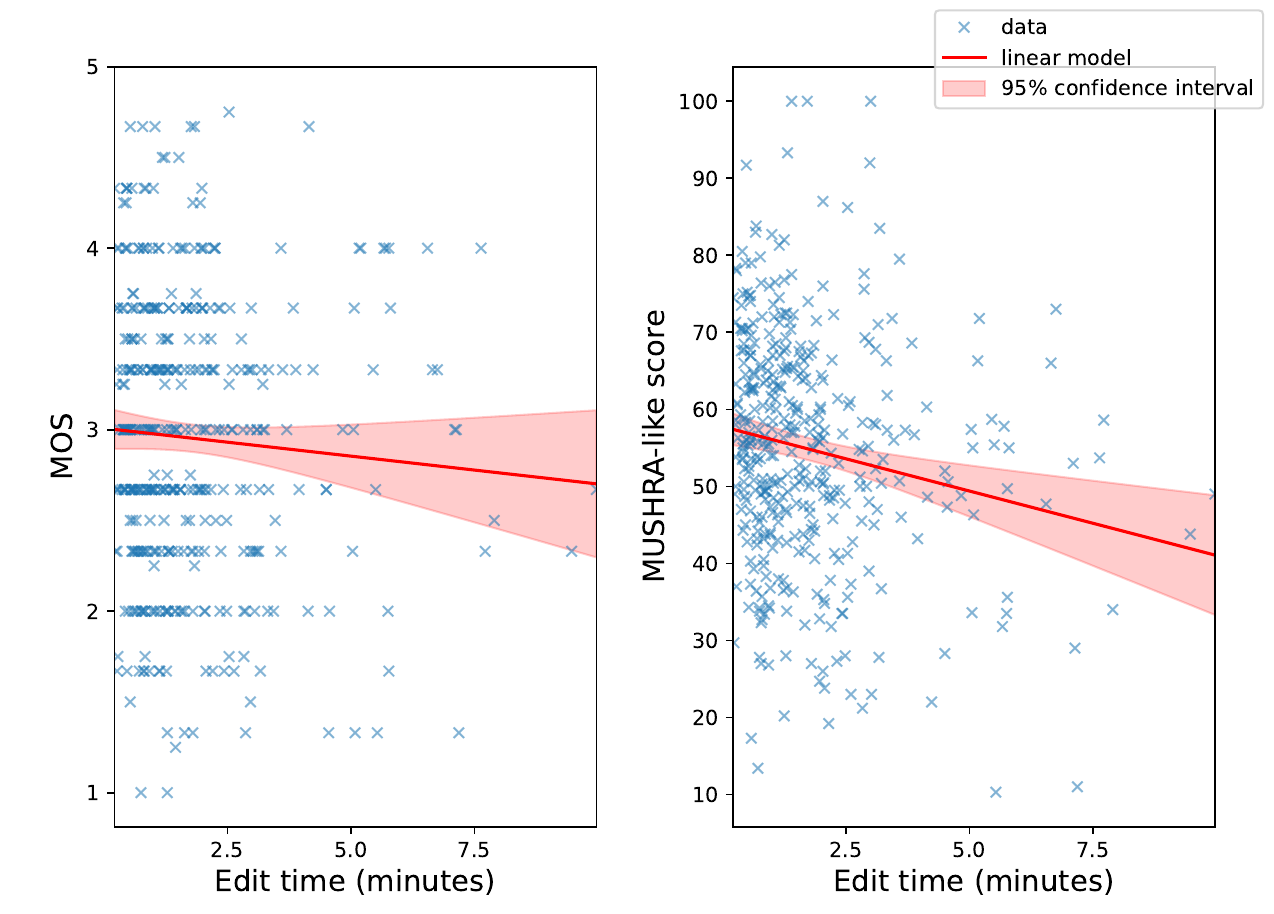}
    \caption{The relationship between HitL effort and the quality of the output. Left: the line of best fit for perceived naturalness as a function of HitL effort, Right: the same but for perceived quality of prosody transfer.}
    \label{fig:effort-vs-quality}
\end{figure}
\vspace{-2mm}

\noindent We finally study the relationship between HitL participant effort and the quality of the output. Here, we take the time spent editing a sample as an indicator of a participant's effort. One might assume that perceived naturalness would be positively correlated with the overall effort of HitL participants. However, this is not the case, as demonstrated by Figure \ref{fig:effort-vs-quality}. We fit a linear regression model to our MOS results as a function of the time taken to make the modifications. Contrary to our assumption, HitL effort negatively correlates with perceived naturalness. There are many plausible reasons for this result, but it underscores the importance of a robust user interface for the proposed method.

\section{Conclusions} \label{sec:conclusions}
Our findings suggest that HitL participants can make meaningful changes to cross-text PT samples. Participants can make the output prosody more appropriate for the target text while preserving the reference prosody. However, this does not lead to an improvement in perceived naturalness. The results shown in Figure \ref{fig:effort-vs-quality} also indicate that more interaction with the model leads to a decrease in naturalness. A notable outcome, highlighted in Figure \ref{fig:distance}, suggests that a reference embedding predicted by a PT model may yield perceptually different prosody, dependent on the target text.

We make simplifying assumptions to reduce overall HitL effort. For example, we do not control pause durations or change word-level \F0 contours in a nuanced manner. Participants therefore had limited control over the full range of prosodic renditions. Despite this, we believe that the current work demonstrates that HitL participants can identify more appropriate prosodic renditions for the cross-text PT task, while preserving the reference prosody. Moreover, our results indicate that under cross-text conditions, \textit{closeness} to the reference embedding is not a reliable metric for measuring prosodic similarity.

\clearpage
\bibliographystyle{IEEEtran}
\bibliography{mybib}

\end{document}